\documentclass[letterpaper, 10 pt, conference]{ieeeconf}  % Comment this line out if you need a4paper
\usepackage[spanish]{babel}
\usepackage[utf8]{inputenc}
\usepackage{listings}% http://ctan.org/pkg/listings
\usepackage{caption}
\usepackage{DejaVuSans}
\usepackage{DejaVuSansMono}
\usepackage{amssymb}
\usepackage{amsmath}

\usepackage{alltt}

\usepackage{fancyvrb}

{\bigskip\VerbatimEnvironment
\begin{BVerbatim}}
{\end{BVerbatim}%
\bigskip}

\usepackage[T1]{fontenc}
\usepackage{fancyvrb}
\usepackage{listings}
\usepackage[scaled=.55]{beramono}    
\fvset{baselinestretch=0.5}
\DeclareCaptionFormat{listing}{\rule{\dimexpr0.9\columnwidth+17pt\relax}{0.35pt}\par\vskip1pt#1#2#3}
\newsavebox{\FVerbBox}
\newenvironment{FVerbatim}
 {\VerbatimEnvironment
  \begin{center}
  \begin{lrbox}{\FVerbBox}
  \begin{BVerbatim}}
 {\end{BVerbatim}
  \end{lrbox}
  \fbox{\usebox{\FVerbBox}}
  \end{center}}

  \usepackage[most]{tcolorbox}
\tcbset{
    frame code={}
    center title,
    left=0pt,
    right=0pt,
    top=0pt,
    bottom=0pt,
    colback=gray!10,
    colframe=white,
    width=0.45\textwidth,
    enlarge left by=0mm,
    boxsep=5pt,
    arc=0pt,outer arc=0pt,
    }

\IEEEoverridecommandlockouts                              % This command is only needed if 
\title{\LARGE \bf
An information model for modular robots: the Hardware Robot Information Model (HRIM)
}

\author{Irati Zamalloa, 
Iñigo Muguruza, \\
Alejandro Hernández,
Risto Kojcev and
Víctor Mayoral\thanks{Erle Robotics} %
}

\usepackage{blindtext}
\usepackage{graphicx}
\usepackage{listings}% http://ctan.org/pkg/listings
\usepackage{xcolor}

\usepackage{hyperref}
\hypersetup{
    colorlinks=true,
    linkcolor=blue,
    filecolor=magenta,      
    urlcolor=cyan,
    citecolor=blue,
}

\usepackage[normalem]{ulem}

\begin{document}
\maketitle
%\tableofcontents
%\thispagestyle{empty}
%\pagestyle{empty}

%%%%%%%%%%%%%%%%%%%%%%%%%%%%%%%%%%%%%%%%%%%%%%%%%%%%%%%%%%%%%%%%%%%%%%%%%%%%%%%%
\begin{abstract}

Today's landscape of robotics is dominated by vertical integration where single vendors develop the final product leading to slow progress, expensive products and customer lock-in. Opposite to this, an horizontal integration would result in a rapid development of cost-effective mass-market products with an additional consumer empowerment. The transition of an industry from vertical integration to horizontal integration is typically catalyzed by de facto industry standards that enable a simplified and seamless integration of products. However, in robotics there is currently no leading candidate for a global plug-and-play standard.

This paper tackles the problem of incompatibility between robot components that hinder the reconfigurability and flexibility demanded by the robotics industry. Particularly, it presents a model to create plug-and-play robot hardware components. Rather than iteratively evolving previous ontologies, our proposed model answers the needs identified by the industry while facilitating interoperability, measurability and comparability of robotics technology. Our approach differs significantly with the ones presented before as it is hardware-oriented and establishes a clear set of actions towards the integration of this model in real environments and with real manufacturers. 

\end{abstract}

%%%%%%%%%%%%%%%%%%%%%%%%%%%%%%%%%%%%%%%%%%%%%%%%%%%%%%%%%%%%%%%%%%%%%%%%%%%%%%%%
\section{Introduction}

Existing and emerging robot technologies have the potential to rapidly disrupt many areas, yet we are still in the early days of the robotics revolution. In 2011, M. Jändel \cite{jandel2011plug} described that one of the main hurdles of the robotics transformation is the lack of integrative standards. According to Jändel, a global standard for plug-and-play robotics would be instrumental for transforming the structure of the industry and dramatically reducing costs, thus facilitating the applications of robotics to all domains of modern society. In summary, a plug-and-play standard that eases interoperability and reusability for robotics would significantly contribute to the development of relevant hardware and software which can quickly be assembled for solving the task at hand.\\

\noindent The importance of reusability and interoperability  in robotics was also highlighted by Mayoral et al. \cite{8046383}, where the authors introduce how the integration effort of a robot, composed by diverse sub-components or parts, supersedes many other  tasks. In that paper, the \emph{Hardware Robot Operating System (H-ROS)} was presented. A common infrastructure that reduces the integration effort by creating an environment where components can simply be connected and interoperate seamlessly. This infrastructure becomes particularly relevant when working with modular robots, robots composed by different sub-modules that may or may not come from the same manufacturer. Seamless and unambiguous communication between robots and their components is a popular topic in the research field of robotics. As introduced in \emph{IEEE Standard Ontologies for Robotics and Automation} \cite{7084073}, similar to humans who require a common and well defined vocabulary for communication, robots present an analogous need. An intermediate standard language with clear and well defined terms is a sine qua non condition for \emph{inter} and \emph{intra} robot interoperability.\\
\newline
Several groups have studied a unified way of representing knowledge and provided a common set of terms and definitions to facilitate interoperability in the robotics domain. Yet, there seem to be conflicting different views on the topic \cite{upm6493}. Leaving aside the somewhat contradictory landscape of definitions and uses of terminology \cite{haidegger2013applied}, our team concludes the following:
\begin{itemize}
    \item Models and ontologies are critical for the development of robotic systems, specially for the interoperability, measurability and comparability of robotics technology.
    \item To the best of our knowledge, most work and studies centered around ontologies for robotics technology are focused on the software abstractions and little discussion has been presented on how this work can be translated to real hardware systems to promote integrability, portability and reusability.
\end{itemize}

\noindent The current state of the art shows that, in an attempt to create standards and international agreements \cite{HAIDEGGER20131215}, several ontologies have been produced in the domain of robotics, however these models still have not been accepted neither translated to industry. Robot component manufacturers still lack of a common set of principles to follow when designing the interfaces of their robot hardware devices. As concluded by Zamalloa et al. \cite{zamalloa2017dissecting}, due to the vertical approach that most robot manufacturers follow and the lack of identified collaborations between them, no single player in robotics has currently the position of establishing a de facto standard by itself.\\

\noindent This work addresses the need in the robotics industry of a common interface that facilitates interoperability among different vendors of robot hardware components. To this end, our work gets inspiration from previous results presented in \cite{8046383, 7084073, DBLP:journals/corr/ZanderHNAEA16, OMG-HAL4RT}, and defines a model for hardware in robotics created through interactions with robot hardware vendors and implemented using the widely spread Robot Operating System (ROS) \cite{quigley2009ros}. In the content that follows, the words \emph{component} and \emph{module} are used interchangeably. Section \ref{related_work} will introduce relevant previous work. Section \ref{HRIM} will introduce the Hardware Robot Information Model (HRIM),  and Section \ref{conclusions} will present conclusions and future work.\\

\section{Related work}
\label{related_work}

\subsection{Ontologies in Robotics}
\label{ontology}

%\todo{Merge this and the following paragraph together}\\
Approaches for ontology-based knowledge engineering in robotics have been studied for years exploring the human-robot and robot-robot interaction. Ontologies sustain not only a common understanding of the domain for humans, but also for robotic systems which need to achieve interoperability, perform their tasks in an autonomous way or interact with humans.\\

%\todo{\textbf{reword this:}
%Ontology-based knowledge engineering approaches are progressively being applied in autonomous robotic systems. Ontologies sustain not only a common understanding of the domain for humans –-necessary for engineering-– but also for human-robot and robot-robot interaction, to represent both the knowledge a robot needs to perform its tasks in an autonomous way or to interact with humans.}\\

\noindent Within \emph{'An IEEE Standard Ontology for Robotics and Automation'} \cite{schlenoff2012ieee}, Schlenoff et al. discuss the development of a standard ontology and its associated methodology for knowledge representation and reasoning in robotics and automation. Such standard aimed to provide a common set of terms and definitions, allowing for unambiguous knowledge transfer among any group of humans, robots, and other artificial systems which were summarized in \cite{7084073}. The purpose of this standard is to provide a methodology for knowledge representation and reasoning in robotics and automation. Open source implementations of it have recently been made available \cite{SandroOntology}.\\

%\todo{be critical about the fact that none of these ontologies/models satisfy the needs of manufacturers. None. Only SW-related aspects.}

%\irati{These two criticisms we can add in a last pharagraph that conclude all the ideas from the related work}

\noindent The W3C Semantic Sensor Networks Incubator Group (SSN-XG)\footnote{\url{http://www.w3.org/2005/Incubator/ssn/}} \cite{j.websem312} aims to build a general and expressive ontology for sensors. According to Compton et al. \cite{Compton09asurvey}, this initiative included members and developers from other ontologies such as CSIRO, MMI and OOTethys. The Open Geo Spatial (OGC) organization adopted the SSN-XG ontology in W3C-OGC project\footnote{\url{https://www.w3.org/TR/vocab-ssn/}} and created SOSA (Sensor, Observation, Sample, and Actuator), a lightweight but self-contained core ontology to cover elementary needs.\\

A recent initiative that used SSN-XG gets described by Zander et al. \cite{DBLP:journals/corr/ZanderHNAEA16}, where they present the \emph{ReApp architecture} that aims to further improve the re-use of robotics software. According to the authors, ReApp builds upon the ROS component model and introduces significant enhancements in terms of metadata and tooling by using a model-driven design methodology backed by a semantic description of software components based on ontologies. ReApp ontologies allow the creation of high-level models of different components and their corresponding interfaces, which can be automatically transformed  to source code and collect relevant information for the developer in a single, integrated step. ReApp was evaluated in a simple industrial scenario with commercially available robots performing different automation tasks. To best of our knowledge, there is no real commercialization nor acceptance of ReApp proposed ontologies in industrial scenarios.

%\todo{critize its hardware relationship. Project didn't continue. Finished more than a year ago. No follow up in the code, nor public uses}.\\

\noindent Space Plug-and-Play Avionics (SPA) \cite{lyke} is a collection of standards designed to facilitate rapid constitution and testing of spacecraft systems using modular components and plug-and-play technology. All the components describe their own functions through and electronic datasheet, by a ontology limited to satellites. From a technical point of view, SPA has the potential to expand into a generic plug-and-play standard for robotics, however, there is no robotics organization supporting it and currently, it is mainly focused on nanosatellites and plug-and-play architectures for space applications \cite{lyke2014lessons}.\\
%\sout{since satellites include sensors, actuation, on board-computing, etc. that is to say they add logic and characteristics by xTEDS (electronic datasheet),}

%\todo{Merge this and the following paragraph together}\\

%\sout{\noindent Although an ontology is critical aspect for the compatibility and the standardization, we can conclude that nowadays there is a lack of a reference ontology that globalizes the needs of robotics industry. Most of the standardization projects for robots have created their own ontology either because its subject was more concrete or by the daily evolution in robotics.}

\noindent In the presented previous work, several attempts have been made to produce standards and international agreements in the robotics domain. However, there has not been wide acceptance of any of the proposed ontologies and there is still a lack of a common set of principles to follow when designing the interfaces of hardware devices. This work focuses on proposing a simple and flexible ontological model that manufacturers of robot hardware components can use to create interoperable devices.

%\sout{Alternatively, HRIM could benefit the current landscape of robotics, satisfying the needs of manufacturers providing them a solution that involve the interoperability in software focused on the hardware. \\}

%\victor{The current state of the art shows that in an attempt to produce standards and international agreements, several ontologies have been produced in the domain of robotics however robot part manufacturers still lack of a common set of principles to follow when designing the interfaces of their hardware devices. Our group identifies the need to provide a simple and flexible model that companies producing robot hardware components can use... }

\subsection{Robotics Models}

\noindent The Object Management Group (OMG\footnote{\url{http://www.omg.org/}}) is an international, open membership, not-for-profit technology standards and industry standards consortium. The work at the OMG is divided in Domain Task Forces (DTF). Particularly, the Robotics DTF, created in 2005, aims to foster the integration of robotics systems from modular components through the adoption of OMG standards. The Robotic Technology Component (RTC) \cite{ando2005rt} \cite{ando2008software}, is one of the standards focused on modular software components created by the OMG. It defines a component model and important infrastructure services applicable to the development of software for robotics. The developers can combine RTCs from multiple vendors into a single application, accelerating the necessary time to create flexible designs. According to the official document, an RTC is a logical representation of a hardware and/or software entity that provides well-known functionality and services. In the proposed documents, the hardware aspect of robotics is not well explained nor defined. The RTC proposed ontological model is discussed in more detail by Stampfer et al. \cite{stampferfiona}. According to them, RTC and SmartSoft were the first available initiatives for specifying  the structures and semantics of a robotic component. The RTC standard had a worldwide impact on the robotics community and raised the awareness about the need of component models and structures for robotics. Yet, the standard itself was not widely accepted in Europe and USA and remained mainly used in Japan.\\

Based on RTC, Japan Embedded Systems Technology Association (JASA) submitted the Hardware Abstraction Layer for Robotics Technology (HAL4RT) to the OMG as a proposed specification. HAL4RT is an Application Program Interface (API) for the layer between an application software of a middleware and the drivers for devices (such as sensor inputs, motor control commands) that increases the portability and reusability of the device drivers. This specification aims to enable device makers, device users and software users to build robotic software without any concern about the differences among the targeted devices. Although, RTC and HAL4RT define a Platform-Independent Model (PIM) \cite{schlenoff2012ieee} for robotic systems, the main objective of each one is different. While RTC standardizes software components in a very general manner, HAL4RT is specifically centered in a subset of existing hardware components.\\

%\sout{\noindent Based on RTC, OMG created JASA submitted HAL4RT to the OMG as a proposed specification, or just say “JASA created the the Hardware Abstraction Layer for Robotics Technology\footnote{\url{http://www.omg.org/spec/HAL4RT/About-HAL4RT/}} (HAL4RT). HAL4RT is an Application Program Interface (API) for the layer between an application software of a middleware and the drivers for devices (such as sensor inputs, motor control commands) that increases the portability and reusability of the device drivers. This specification aims to enable device makers, device users and software users to build robotic software without any concern about the differences among the targeted devices. Although, RTC and HAL4RT define a Platform-Independent Model (PIM) \cite{schlenoff2012ieee} for robotic systems, the main objective of each one is different. While RTC standardizes software components in a very general manner, HAL4RT is specifically centered in a subset of existing hardware components.}\\

However, it seems that HAL4RT has not gone beyond 1.0 beta version. We must highlight that, apart of HAL4RT 1.0\footnote{\url{http://www.omg.org/spec/HAL4RT/About-HAL4RT/}} document, the rest of the remaining information is only available in Japanese. For example, Open Embedded Library (OpenEL) 3.0, an implementation of HAL4RT developed by Upwind Technologies, is described entirely in Japanese. 

\section{Hardware Robot Information Model (HRIM)}
\label{HRIM}

In this section we present HRIM, a common interface that facilitates interoperability among different vendors of robot hardware components with the purpose of building modular robots. HRIM focuses on the standardization of the logical interfaces between robot modules, designing a set of rules that each device has to meet in order to achieve interoperability. \\

% \todo{rewrite!!}
% \todo{Nowadays, the importance of reusability and interoparability in robotics, that many companies and research centers seek a short-term solution. And, the modular robots has the demanded flexibility in creating and adapting robots to customize the process, that is closely connected with the standardization, since it reduces the robot building process to the half thanks to removing all the hardware integration effort. HRIM focus the effort in the hardware solving one of the biggest problem of robotics: the incompatibility in terms of communication between components. So, the modular robots are built with modular hardware which in turn is achieved with a modular interface: HRIM.\\}

Even though HRIM\footnote{\url{https://therobotmodel.com/}} was born as part of the H-ROS project, it is an independent standard interface for robot modules which contains rules/specifications that standardize interactions between different robot components. Similar to other robotics standardization approaches (such as ReApp or BRICS), HRIM builds upon the component model of ROS, since it stands out as one of the largest integration platforms with implementations and mappings to several languages and platforms. The popularity of ROS led to a huge variety of new algorithms and solutions of technical challenges in robotics. ROS is a representative example of the current situation in robotics software. Already in 2009, it was mentioned as the most promising emerging standard in the Roadmap for US Robotics \cite{hollerbach2009roadmap}, and since then, it has only grown enabling the reuse of the code and creating simulation platforms to support early development and testing of algorithms, without compromising the safety of researchers and hardware.\\

%to achieve the interoperability between components even though they are from different vendors of robot (hardware) components. 

HRIM aims to complement ROS with a standardization effort focused on hardware. It is a model that defines the software interface which the different robot components have to meet in order interoperate seamlessly. HRIM is designed to be implemented by companies which are manufacturing modular hardware components, in order to facilitate the integration effort when building robots.\\

HRIM shares some correlation with different standardization projects, specifically with HAL4RT and OpenEL presented in 2012\footnote{\url{https://staff.aist.go.jp/t.kotoku/omg/2012Burlingame/robotics2012-12-15.pdf}}, where they proposed an open platform to standardize the specifications of the software implementation of robotics and control systems unifying the different interfaces of each device manufacturer with the objective of enable applications running on different hardware. However, we must highlight some differences and enhancements:

%This increases the portability and reusability of the software, resulting in improved quality, lower costs and productivity lead to improved productivity offering great advances for users and developers. 
%However, we must highlight some differences and enhancements:

%HRIM shares some correlation with different standardization projects, specifically with HAL4RT 2.0 \cite{OMG-HAL4RT} from OMG. Although the objective of HAL4RT is different, this project is the closest to our vision. However, we must highlight some differences and enhancements:

%H-ROS, The Hardware Robot Operating System, delivers a solution for companies manufacturing sensors, actuators, etc. to create a modular robot components that interoperate and can be easily reused; even from different manufacturers, you will only have to connect them together. With H-ROS you can create plug and play devices simplifying the robot building process.

\begin{itemize}
    \item HRIM considers all types of devices necessary for the construction of a robot including but not limited to components that provide: \emph{sensing}, \emph{actuation}, \emph{communication}, \emph{cognition}, \emph{user interfaces} or \emph{power}.
    %and \emph{composite}. %While HAL4RT only specifies the first two.
    %\item HRIM is based on the ROS component model, that is, it adopts most of its abstractions, while HAL4RT defines a Platform-Independet Model (PIM).
    \item HRIM infrastructure is based on a thorough market analysis of each type of device. These interactions with manufacturers allow the model to take into account all the possibilities (features) that the robotic hardware market offers.
    %\item HRIM aims to include relevant and related hardware standardization efforts such as HAL4RT. The ROS component model is, in some aspects, already compatible with HAL4RT. Therefore, HRIM is HAL4RT compliant. Details about its conformance are covered in Appendix \ref{conformance}.
    \item HRIM defines the units used for all the devices, taking into account the official repository of ROS \cite{standard_ROS} and the International System of Units \cite{standard_SI}, enabling a well defined and correct communication between devices and systems.% In this way the user can replace one device with another (whether or not the same manufacturer) and it will start working automatically since it comprises the code made for the previous one in its entirety.
    %For example, they standardize the motors with position control, velocity control and torque control, but what is happen with the components that has the possibility to change the acceleration? For the other hand, there are servos that has the possibility to work with 360 degrees, but the are others that work only in an specific area. The user would be able to know this kind of critics characteristics easily without having to worry searching the datasheet. For that we have include the Specs inside all of the devices.
    \item HRIM contemplates all the robotics hardware, so that users can build any type of robot. For example, in the Actuator category, besides servomotors, different types and subtypes of actuation devices are being defined. Such as a gripper, a vacuum gripper, etc.
    %\item As we have read in the HAL4RT 2.0 delivery, it has made based on the implementation of OpenEL (Open Embedded Library) 3.0 by JASA (Japan Embedded Systems Technology Association). We have been trying to learn and read more about it and has been imposible since the most of the documentation and web pages are written in Japanese. Apart of that it sound that they have done the modifications on the standard to be complied with their work, honestly it sounds bad, the correct way to do it is modify the OpenEL to comply with the standard.
    %\item We are not agree with the definition on Actuator that HAL4RT propose \emph{The Actuator element defines an API that the Actuator device with one degree of freedom should have. } As we have understood, it is only for servomotors (rotary and linear). We identify more devices inside the Actuators, for example, gripper, vacum or speaker.
    \item Compared with HAL4RT and OpenEL, HRIM takes into account devices with complex data streams. For instance, a camera can send an image or a video.
\end{itemize}

Comparing to other related work and standards, HRIM aims for generality, which is essential for acceptance of the model and for achieving global hardware standardization of robotic components.\\ 

%\risto{I dont think this paragraph is necessary. It brakes the flow and is unconnected to previous text, maybe it fits better in the description of HAL4RT in related work or something...}\todo{\noident We must highlight that apart of the OMG official document the rest of the remaining information is only available in Japanese, for example Open Embedded Library (OpenEL) 3.0, an implementation of HAL4RT developed by Japan Embedded Systems Technology Association (JASA), is described entirely in Japanese. In addition, a curious detail that must be mentioned is that OpenEL has conditioned all the modifications to which the standardization has been submitted in its 2.0 version, which makes us think that the only objective of JASA is that its own technology is backed by a standardization}.\\

The following subsections are focused on explaining the concepts of HRIM and its potential benefits for applicability in robotics. In section \ref{HRIM_classification} the classification of hardware modules is explained. Section \ref{HRIM_naming} introduces the naming convention followed within HRIM. Section \ref{HRIM_structure} describes the HRIM general structure. This explanation has been complemented with section \ref{HRIM_example}, where an HRIM component model example is provided: the model of a servomotor module, which helps understanding the whole explained theory in a real context.

% Removed for now.
%\todo{With the Section \ref{HRIM_simulation} we delve into simulation, what unites the modular robots with the Artificial Intelligence technologies.} 

\subsection{Module classification}
\label{HRIM_classification}

In order to build any robot, real world implementation requires taking into account the common hardware components used in robotics. The robot modules have been classified in 6 types of modules which correspond to the task they can perform: 
%Although the classification is done for H-ROS infrastructure, it is a good form to ordering all the devices (bio-inspired structure) and to simplify the interaction:
%HRIM want to standardize most of the (hardware) components used to build a robot. The robot modules have been classified in 7 types, having into account device general purpose. Although the classification is done for H-ROS infrastructure, it is a good form to ordering all the devices (bio-inspired structure) and to simplify the interaction:\\
\begin{itemize}
    \item \textbf{Sensors:} help robots perceive its environment and share the information with the rest of the connected modules or with the user.
    \item \textbf{Actuators:} components within a robot that provide means of physical interaction with the environment.
    \item \textbf{Communication:} provide means of interconnection between the modules of the robot or expose new communication channels to the overall network.
    \item \textbf{Cognition:} specialized in computation and coordination, these modules perform most of the computationally expensive tasks within the robot. %It also deals with the reconfiguration process of H-ROS parts (plug and play functionality) and exposes ROS interfaces.
    \item \textbf{User Interfaces (UI):} provide means of interfacing with the robot, typically related to human-robot interaction such as joysticks, tactile screens or voice input.
    \item \textbf{Power:} components whose purpose is to deliver power to the system or subsystems.
    %\item \textbf{Composite:} sophisticated sub-systems formed by composing basic components among the previous classes to achieve a totally different purpose.\\
\end{itemize}
Each type is composed by sub-types or \textbf{devices} related to the functionality of the component. For example, a camera is a sub-type of the sensor type.

\subsection{Naming convention}
\label{HRIM_naming}
The standardization of the naming rules listed, allows the hardware vendor, developer or robot operator (user) to facilitate interoperability and reduce the time to read and understand the code, focusing on more important tasks than the syntax. The naming convention presented below is based in the ROS component model:\\  %So, the naming convention simplifies the work of all the HRIM users standardizing the most important ROS 2\footnote{\url{https://wiki.ros.org/ROS/Patterns/Communication}} concepts thank to the representation listed in Table \ref{naming_convention}.\\
\begin{tcolorbox}
%\textbf{\small Naming convention}\\
\centering
\textbf{Package}
\begin{FVerbatim}
hrim_<device_kind>_<device_name>_<vendor_id>_<product_id>
\end{FVerbatim}
\textbf{Node}
\begin{FVerbatim}
hrim_<device_kind>_<device_name>_<instance_id>
\end{FVerbatim}
\textbf{Topic}
\begin{FVerbatim}
hrim_<device_kind>_<device_name>_<instance_id>/<topic_name>
\end{FVerbatim}
\textbf{Message}
\begin{FVerbatim}
hrim_<device_kind>_<device_name>_msgs/msg/<message_name>.msg
\end{FVerbatim}
\textbf{Generic Message}
\begin{FVerbatim}
hrim_generic_msgs/msg/<message_name>.msg
\end{FVerbatim}
\textbf{Service} 
\begin{FVerbatim}
hrim_<device_kind>_<device_name>_<instance_id>/<service_name>
\end{FVerbatim}
\textbf{Service} (file)
\begin{FVerbatim}
hrim_<device_kind>_<device_name>_srvs/srv/<service_name>.srv 
\end{FVerbatim}
\textbf{Action}
\begin{FVerbatim}
hrim_<device_kind>_<device_name>_action/<action_name>.action
\end{FVerbatim}
\textbf{Parameter}
\begin{FVerbatim}
param name="<parameter_name>" type="<data_type>
value="<parameter_value>"
\end{FVerbatim}
\end{tcolorbox}

\noindent where:

\begin{itemize}
    \item \verb|hrim|:  Hardware Robot Information Model. 
    \item \verb|<device_kind>|: module classification, explained in section \ref{HRIM_classification}.
    \item \verb|<device_name>|: or sub-type, for example, camera.
    \item \verb|<vendor_id>|: identifier for the device vendor. %\todo{think about how we can do it. numbers, numers+letters etc.}
    \item \verb|<product_id>|: identifier for the product, typically, the manufacturer part-number. %The device vendor gives a unique numbering. %In the previous HAL4RT document they do like this: An Product ID is defined as a 32bit unsigned interger type, with values between 0x00000000 and 0xFFFFFFFF. The instance IDs are defined by the application developers. They are defined by device suppliers and manufacturers.
    \item \verb|<instance_id>|: unique identifier for each device. %When multiple products of the same type are used in the targed system, they are used to identify them.%\todo{think about how we can do it. numbers, numers+letters etc.} %The application creator sets in advance by some means.%In the previous HAL4RT document they do like this: An Instance ID is defined as a 32bit unsigned interger type, with values between 0x00000000 and 0xFFFFFFFF. The instance IDs are defined by the application developers.
    \item \verb|<topic_name>|: descriptive term for each topic.
    \item \verb|<message_name>|: descriptive term of each message.
    \item \verb|<parameter_name>|: descriptive term of each parameter.
    \item \verb|<data_type>|: type of data for the corresponding parameter.
    \item \verb|<parameter_value>|: the value contained in the parameter.
\end{itemize}

\begin{figure*}[!htb]
\centering
\includegraphics[width=\textwidth]{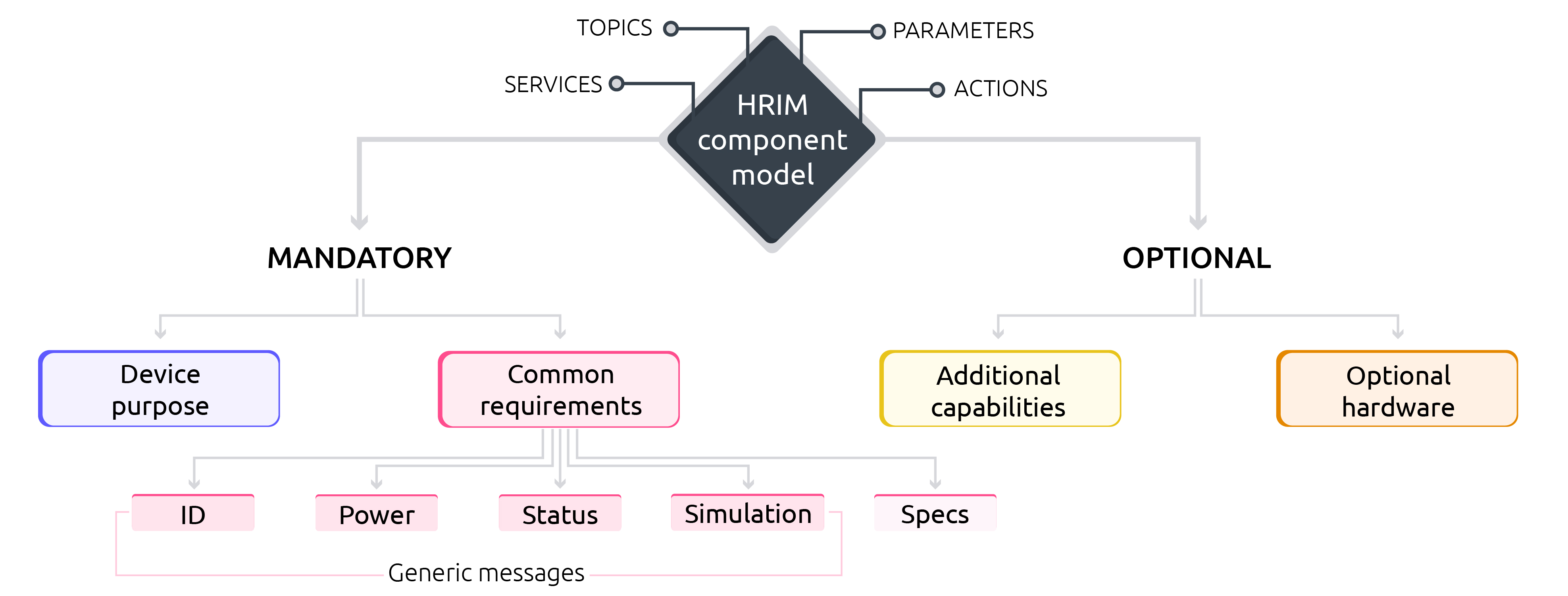}
\caption{\small The general structure in which all the HRIM component models are based on. Each component has \emph{topics}, \emph{services}, \emph{parameters} and \emph{actions} to communicate. For each one of these abstractions, the figure illustrates that some will be mandatory and some others optional.}
\label{general_structure}
\end{figure*}

\subsection{General structure}
\label{HRIM_structure}

The HRIM information model consists of different HRIM component models as illustrated in Figure \ref{general_structure}, each representing a different device sub-type used in modular robots. These component models are built with the ROS communication abstractions: topics, services, parameters and actions. All of them are labeled as mandatory or optional, in order to inform the device manufacturer if the information must be included or not. The optional aspects are related to the characteristics of each particular device. In the following paragraphs, the two aspects are explained in detail to understand the difference and the correct use of both of them. \\

%The first criterion that must be clear is the obligation with which all the details are labeled: MANDATORY (needed information, the device vendor has the obligation to include) and the OPTIONAL (extra characteristics, depending the features of the device can be included or not).  \\

%Although the construction of HRIM involves many things that have to be fitted in a general plane, the components must be analyzed one by one in order to define them in detail. What leads to a thorough analysis of each of it to identify the key features on the one hand and also define all the possibilities that the component can adopt on the other hand. So, the most of the manufacturers will be able to standardize their device taking into account all the features they offers.\\

%All the modules will follow the same structure, synthesized in Figure \ref{general_structure} to achieve a standardization that can be easily implemented by any hardware manufacturer in a simple way.\\

%The content, as we have mentioned before is built on top of ROS using ROS syntax: topics, messages, services etc. 

\subsubsection{\textbf{Mandatory}} is the content that all the modules must include, in order to enable interoperability between different components. \\ 

\begin{itemize}
    \item \textbf{Device purpose}: HRIM detects the essence of the device and captures this information in the \emph{device purpose} abstractions. These elements define the basic information that allows the user to work with the component. It is a customized information for each sub-type of module that globalizes to all of its kind, making them interoperate, even when coming from different manufacturers. The \emph{device purpose} of the module is what makes it different from other sub-types, so it has to be composed by at least a topic, a service, an action, or a mix between them, complemented by parameters.\\
    
    \item \textbf{Common requirements}: capture various information to improve the user experience and ensure a correct operation of the whole robotic system. As pictured in Figure \ref{general_structure}, four of them use generic messages such as: \emph{ID.msg}, which publishes the general identity of the component, \emph{Power.msg}, that publishes the power consumption, \emph{Status.msg}, which inform about the resources that are consumed, \emph{Simulation3D.msg} and \emph{SimulationURDF.msg}, that sends the device 3D model and related information. \emph{Specs<DeviceName>.msg} is a custom message which reports the main features of the device. \\  
    
\end{itemize}

%\begin{figure}[h!]
%\centering
% \includegraphics[width=0.4\textwidth]{HRIM_mandatory4.jpg}
%\caption{\small Mandatory information}
%\label{mandatory}
%\end{figure}
\subsubsection{\textbf{Optional}} these are additional capabilities which will be included depending on the particular characteristics of each device. The manufacturer of each component will be in charge of including the optional information or not. As showed in Figure \ref{general_structure}, there are two groups, additional capabilities and optional hardware, defined in order to simplify the whole HRIM, while respecting the modularity mentioned at the beginning of this section.\\  

\begin{itemize}
    \item \textbf{Additional capabilities}: capture complementary aspects to the component \emph{device purpose}. Usually these are topics, services, actions or parameters customized for each component sub-type. For example, a camera able to control the brightness of the image needs a parameter to adjust such, therefore it is categorized as optional due to the fact that not all cameras contain such capability.\\
    
    \item \textbf{Optional hardware}: many devices include additional sub-devices enhancing their own capabilities. \emph{Optional hardware} capture these aspects. For example, a camera device could include a microphone. In that case, the content of the HRIM michophone component model would be added as \emph{optional hardware} within the HRIM camera component model. 
\end{itemize}

\begin{figure}[h!]
\centering
 \includegraphics[width=0.48\textwidth]{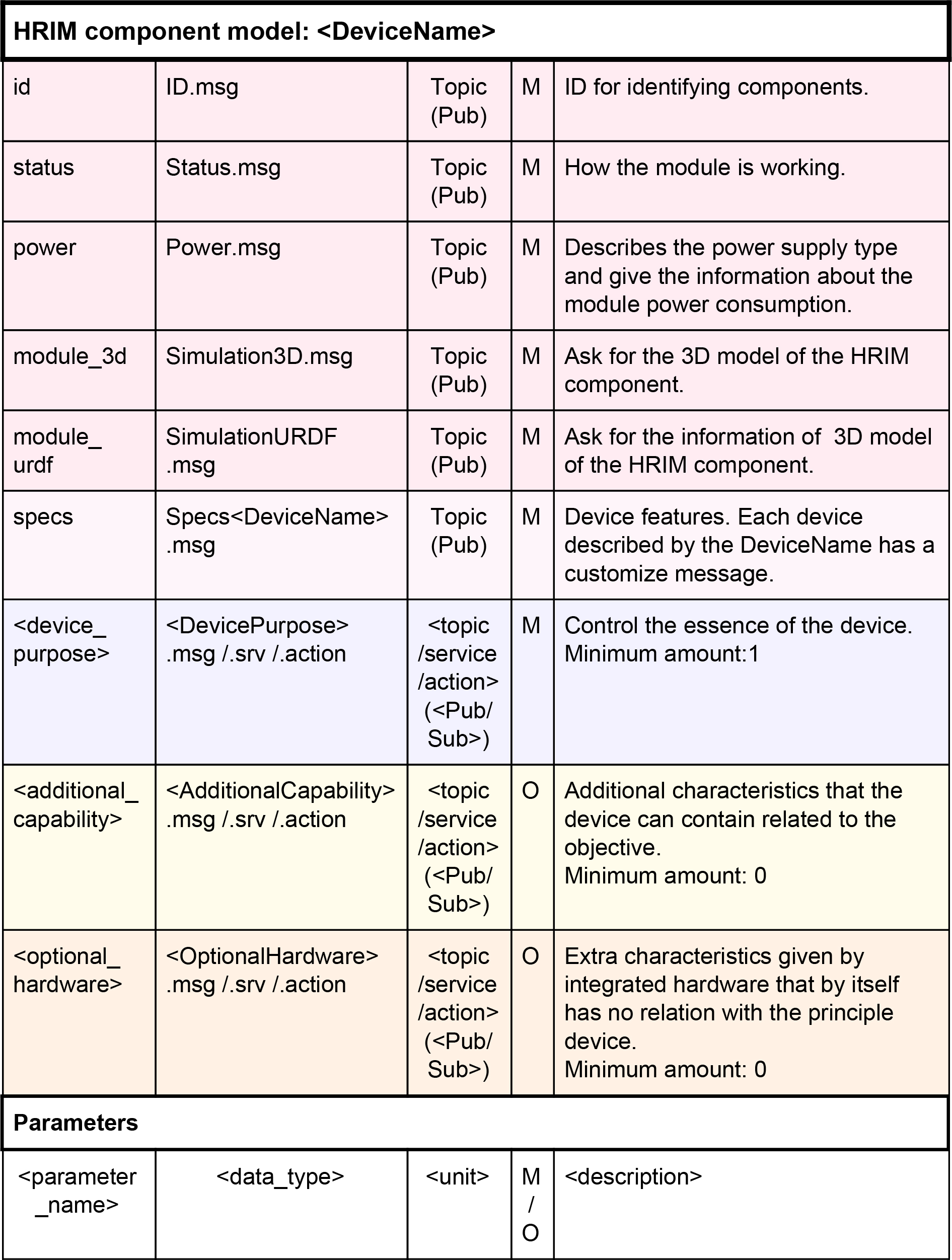}
\caption{\small HRIM component model summary including the common aspects.}
\label{common}
\end{figure}

\begin{figure}[h!]
\centering
\includegraphics[width=0.48\textwidth]{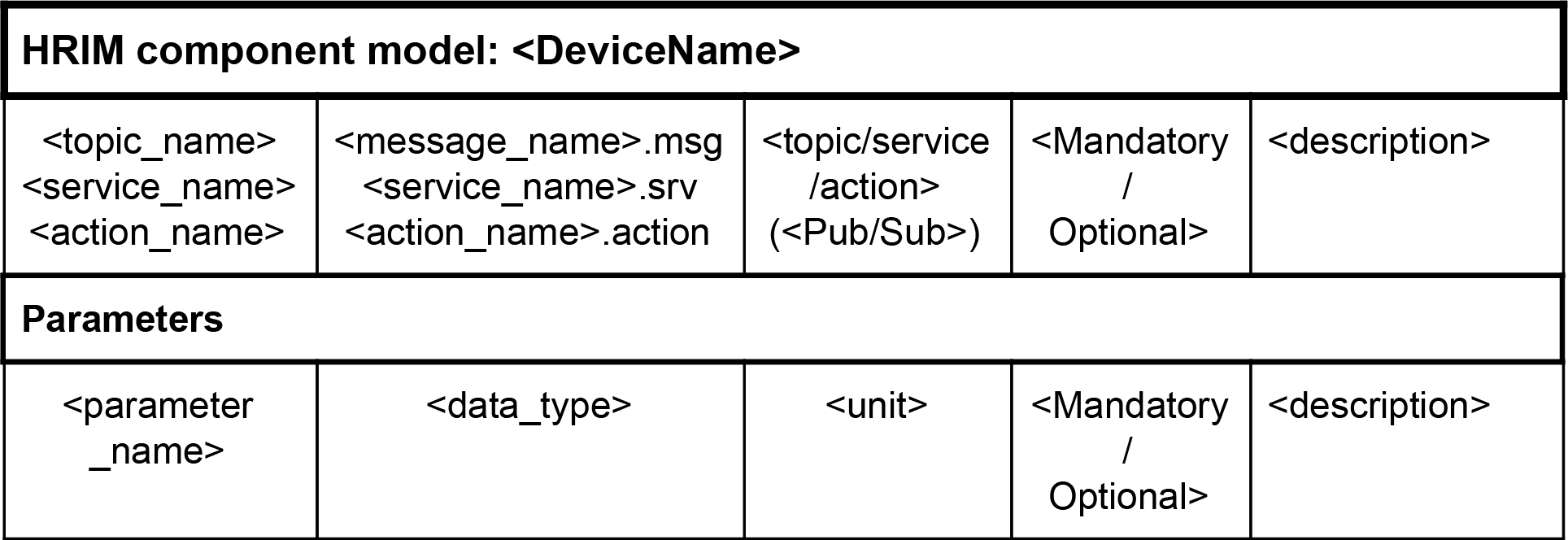}
\caption{\small HRIM component model summary template. On the top, the name of the device that is summarized is showed, then the \emph{topic}, \emph{service} and \emph{action} name with the corresponding \emph{.msg}, \emph{.srv} or \emph{.action} and the direction (if it is needed): Pub:published or Sub:subscribed. The parameters are described by the type of data and its corresponding unit.}
\label{template}
\end{figure}

Each HRIM component model will be summarized through the template showed in Figure \ref{template}, which is divided in two main categories: the first category includes the \emph{topics}, \emph{services}, \emph{actions} and the second, the \emph{parameters}, all of them labeled with mandatory (M) or optional (O). Although the HRIM component model is created for each device, all of them follow the same general structure with some common aspects illustrated in Figure \ref{common}. Furthermore, the color code used in Figure \ref{general_structure} has been respected. \\

%As is mentioned before, there are some generic topics that all the modules has to contain which the Figure \ref{common} shows in green color. that all the modules has in common. The Figure \ref{common}, shows the appearance that each component will follow, we have marked with color to be more visual and more understandable, for example the green color represent the information that all components has to publish trough the ROS messages. 

The first five topics listed in Figure \ref{common} are the common topics that use generic messages.
%The extended version of the common topics are presented in Appendix \ref{msg}:\\
%The topics marked with green color in Figure \ref{common} represents the generic hrim information which will use the same messages: (although the following description visualize a summarized version of the messages, it is available the whole content in Appendix \ref{msg}).\\
\begin{itemize}
    \item \textbf{id:} \emph{ID.msg} is the identification of each module. It is designed to inform the user and also the system about the component itself. \emph{/id} topic should be programmed to be released only when the user requires it.\\

% \begin{tcolorbox}[colback=magenta!10]
% \begin{FVerbatim}
% hrim_generic_msgs/msg/ID.msg
% \end{FVerbatim}

% \begin{lstlisting}
% unit8 device_kind_id     # device classification
% string device_name       # sub-type, for example, camera
% uint32 vendor_id         # identifier of the vendor
% unit32 product_id        # part-number
% unit32 instance_id       # identifier of the module
% ...
% \end{lstlisting}
% \end{tcolorbox}

\end{itemize} 

\begin{itemize}
    \item \textbf{power:} \emph{Power.msg} publishes information of the power source (power supply or PoE), and the current power consumption. The \emph{Power.msg} provides continuous information regarding the power consummation and power requirements of each module, allowing the easy construction of modular robots.

% \begin{tcolorbox}[colback=magenta!10]
% \begin{FVerbatim}
% hrim_generic_msgs/msg/Power.msg
% \end{FVerbatim}
% \begin{lstlisting}
% float32 poe_voltage
% float32 poe_current
% float32 poe_power

% float32 supply_voltage
% float32 supply_current_consumption
% ...
% \end{lstlisting}
% \end{tcolorbox}
\end{itemize}

\begin{itemize}
    \item \textbf{status:} \emph{Status.msg} is designed to inform about the resources that are being used by each component. Similar to the \emph{Power.msg}, it is programmed to publish continuously in order to simplify the robot building process.\\

% \begin{tcolorbox}[colback=magenta!10]
% \begin{FVerbatim}
% hrim_generic_msgs/msg/Status.msg
% \end{FVerbatim}
% \begin{lstlisting}
% float32 system_cpu		    # Total CPU utilization in %
% float32 core_temperature  # CPU temperature in Celsius
% uint32 system_ram_total	  # Total RAM in MB (megabyte)
% uint32 system_ram_used	  # Used RAM in MB (megabyte)
% ...
% \end{lstlisting}
% \end{tcolorbox}
\end{itemize}

\begin{itemize}
    \item \textbf{simulation:} \emph{Simulation3D.msg} and  \emph{SimulationURDF.msg} allow the user to obtain the 3D model and a URDF fragment of each module in order to build a virtual robot which then can be easily used in simulation frameworks, such as Gazebo or MoveIt. The simulation topics do not start publishing until they detect a user interface or a simulation framework connected to the system. %\todo{Section \ref{HRIM_simulation} describes in more details the simulation and visualization of modular components.} %Remove if we finally decide not to delve into the simutation in the Section E.

% \begin{tcolorbox}[colback=magenta!10]
% \begin{FVerbatim}
% hrim_generic_msgs/msg/Simulation3D.msg
% \end{FVerbatim}
% \begin{lstlisting}
% byte[] 3d_model		    # .stl of the component
% \end{lstlisting}
% \end{tcolorbox}

% \begin{tcolorbox}[colback=magenta!10]
% \begin{FVerbatim}
% hrim_generic_msgs/msg/SimulationURDF.msg
% \end{FVerbatim}
% \begin{lstlisting}
% string urdf_model	    # the urdf corresponding to the .stl
% \end{lstlisting}
% \end{tcolorbox}
\end{itemize}

The rest of the HRIM component model abstractions, such as \emph{specs} or \emph{<device\_purpose>}, are specifically designed and customized for each component. To give a better understanding of these concepts, a practical example is given in subsection \ref{HRIM_example}.

%\irati{To comply with HAL4RT take this into account:}
%\todo{The new tool that is expected from Open Robotics, they are working on it, is the Supervisor tool. Is like an auditor that control all is working property: the node are correcting in the correct way and also the times are respecting. If not, Supervisor tool create an event in order to inform what the event type and the error type is. With this we are complying with the HALObserver and with the TimerObserver that HAL4RT specifies. \todo{HAL4RT, in the document that I have, doen not specifies the event types and also the errors.}}\\

%\todo{The ReturnCode that the HAL4RT is talking about is something internal of the component. To comply with this we have to implement this in the code. Is not a new .msg or new .srv. We will create a list of error that when the code is executing the user will know what is happen. For example, power error. \todo{create a list of errors and 1 notification has to be that all is working OK.}}\\

%\todo{The operatins Int, ReInt, Finalize that all the HALComponent must include we can implement using ROS Lifecycle. It is more complete that the one that HAL4RT define: State transition.}\\

%\todo{Property, a programming thing, easy to add something like this in order to comply with HAL4RT}\\

\subsection{HRIM component model example: rotary servomotor}
\label{HRIM_example}
This section elaborates the practicality of the HRIM model trough its implementation on a real device. We have chosen the rotary servomotor since it is one of the most used components in robotics and sufficient to explain most of the HRIM details.\\ 

HRIM defines a rotary servomotor as a smart actuator that creates a circular movement and allows for precise control of position, velocity, effort and, sometimes, acceleration. Apart from the common requirements (ID, Status, Power, Specs and Simulation, detailed in the previous section), all rotary servomotors will contain \textit{a topic referring to the circular movement}. Figure \ref{example_servo} presents a summary of the HRIM rotary servomotor model on which the manufacturers will have to base themselves to configure their own rotary servomotor. The hardware maker must include, at least, the mandatory aspects. The optional parameters can be added if the module contains these additional features.\\
%\sout{All the rotary servomotors will contain \textbf{a topic referring to the circular movement} in order to cover the rotary servomotors purpose --apart from the common requirements: ID, Status, Power, Specs and Simulation, detailed in the previous Section.\\}

The following paragraphs provide a walkthrough of the abstractions proposed for the HRIM component model of a rotary servomotor:

\begin{figure}[h!]
\centering
\includegraphics[width=0.48\textwidth]{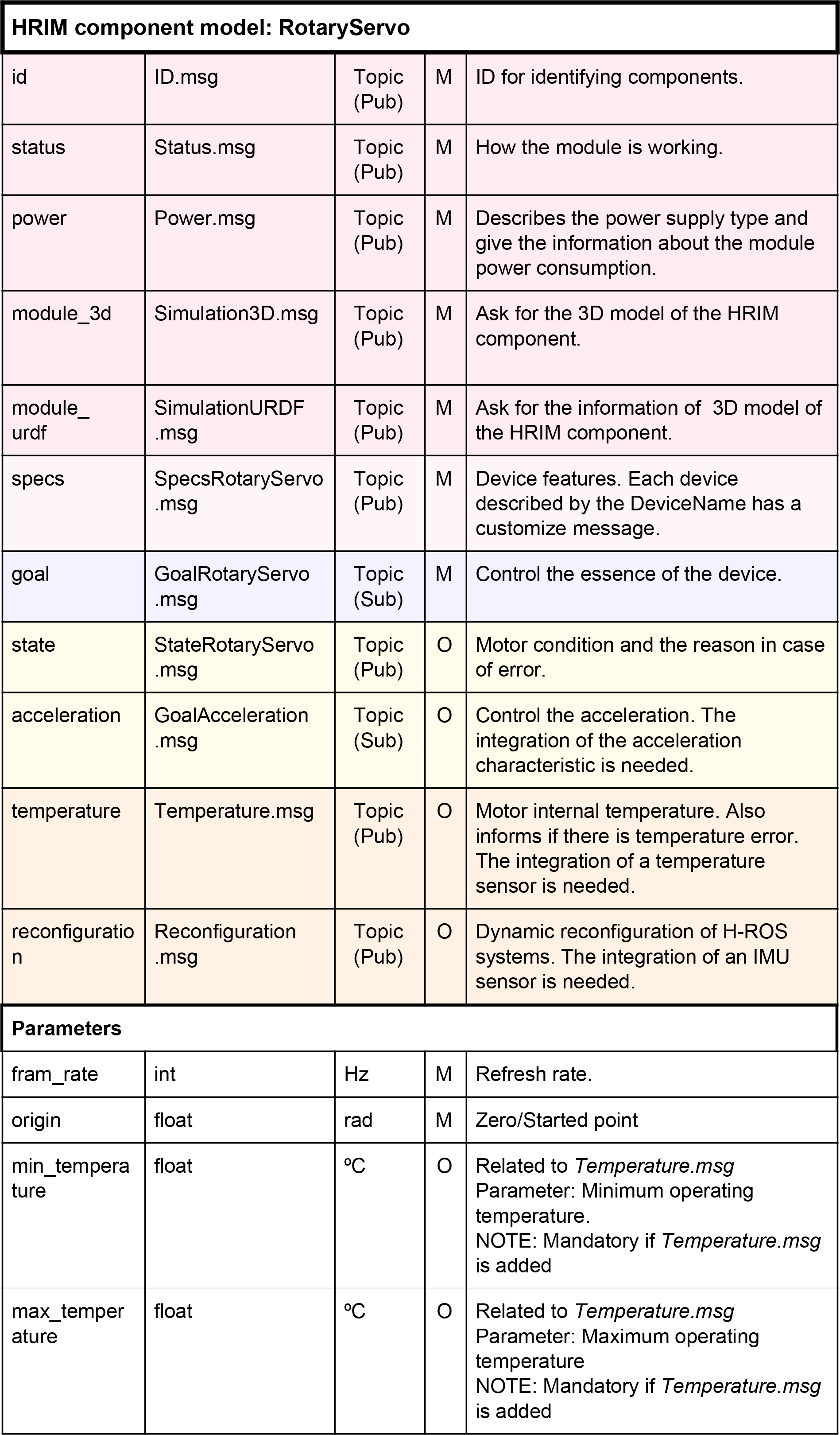}
\caption{\small Rotary servomotors model summary.}
\label{example_servo}
\end{figure}

\begin{itemize}
    \item \textbf{Specs:} \emph{SpecsRotaryServo.msg} describes all the necessary specifications for the implementation of a rotary servomotor.\\

% \begin{tcolorbox}[colback=magenta!5]
% \begin{FVerbatim}
% hrim_actuator_rotaryservo_msgs/msg/SpecsRotaryServo.msg
% \end{FVerbatim}
% \begin{lstlisting}
% uint8 control_type       # servomotor control type
% float64 range_min        # servomotor work range max (rad)
% float64 range_max        # servomotor work range min (rad)
% float64 precision        # angular precision (rad)
% ...
% \end{lstlisting}
% \end{tcolorbox}
\end{itemize}

\begin{itemize}
    \item \textbf{Device purpose:} rotary servomotors belong to the actuator type classification. In this case, it is necessary to define an end goal (position) which can be achieved with different velocity or acceleration. \\

% \begin{tcolorbox}[colback=blue!5]
% \begin{FVerbatim}
% hrim_actuator_rotaryservo_msgs/msg/GoalRotaryServo.msg
% \end{FVerbatim}
% \begin{lstlisting}
% uint8 control_type   # servomotor control type

% float64 position     # Position to move (rad)
% float32 velocity     # Velocity to move (rad/s)
% float32 effort       # Force to move (kg*cm3)

% \end{lstlisting}
% \end{tcolorbox}
\end{itemize}

%To check if the servomotor has any additional features, the user can call common ROS commands such as: \emph{rostopic list}, \emph{rosparam list}, \emph{rosservice list} and \emph{rosaction list}. 
Servomotors always contain the common information defined for the actuator category. In the rotary servomotor case, it has four optional topics. The first two are \textbf{additional capabilities} that complement the rotary servomotor purpose. The rest are independent sensors that can appear integrated in the component, called \textbf{optional hardware}.\\

\begin{itemize}
    \item \textbf{State:} through \emph{StateRotaryServo.msg}, a rotary servomotor continuously publishes information regarding the motor condition, how is it working, and the reason in case of error.\\
    %\todo{To include this topics it is neccesary, maybe y have to split this message in differents dependig the included sensors.} 
    
% \begin{tcolorbox}[colback=yellow!5]
% \begin{FVerbatim}
% hrim_actuator_rotaryservo_msgs/msg/StateRotaryServo.msg
% \end{FVerbatim}
% \begin{lstlisting}
% float64 goal          # commanded position (rad)
% float64 position      # current position encoder (rad)
% float64 error         # difference between current-goal position (rad)
% float64 velocity      # current velocity (rad/s)
% ...
% \end{lstlisting}
% \end{tcolorbox}

    \item \textbf{Acceleration:} some rotary servomotors have the option to control the acceleration. If that is the case, the manufacturer has to use the \emph{GoalAcceleration.msg} and expose the \emph{acceleration}, so that the user can control this feature.\\
    
% \begin{tcolorbox}[colback=yellow!5]
% \begin{FVerbatim}
% hrim_actuation_rotaryservo_msgs/msg/GoalAccleration.msg
% \end{FVerbatim}
% \begin{lstlisting}
% float32 acceleration   # angular acceleration (rad/s2)
% \end{lstlisting}
% \end{tcolorbox}

    \item \textbf{Temperature:} the temperature sensor is an independent sensor which can usually be integrated into other devices in order to know the interior temperature of the module. As you can see in Figure \ref{example_servo}, this topic goes hand in hand with two parameters that, in this case, are mandatory to be able to control the temperature in a correct way.\\
% \begin{tcolorbox}[colback=orange!10]
% \begin{FVerbatim}
% hrim_sensor_thermometer_msgs/msg/Temperature.msg
% \end{FVerbatim}
% \begin{lstlisting}
% float64 temperature         # Current tempreture in Celsius
% bool    temperature_error
% \end{lstlisting}
% \end{tcolorbox}
    
    \item \textbf{Reconfiguration:} Through messages like \emph{Reconfiguration.msg}, a module is able to inform about its particularities, so that it can be automatically integrated in the robot with as little interaction from humans as possible. This reconfiguration functionality is expected to be extended in the future, and likely, additional types of reconfiguration will be included within HRIM.\\
    
    %in order to use this property. To achieve reconfigurability each module contains IMU sensor.
    %H-ROS has patented the automatic reconfigurability for robots. So, all the H-ROS compliant modules has to include \emph{Reconfiguration.msg} in order to use this property. To achieve reconfigurability each module contains IMU sensor.

% \begin{tcolorbox}[colback=orange!10]
% \begin{FVerbatim}
% hrim_hros_msgs/msg/Reconfiguration.msg
% \end{FVerbatim}
% \begin{lstlisting}[belowskip=-0.8 \baselineskip]
% geometry_msgs/Quaternion orientation
% float64[9] orientation_covariance # Row major about x, y, z axes
% geometry_msgs/Vector3 angular_velocity
% float64[9] angular_velocity_covariance # Row major about x, y, z axes
% ...
% \end{lstlisting}
% \end{tcolorbox}
\end{itemize}

As an information model for modular robots, HRIM has been built with modularity in mind. Each block has a unique and well identified purpose, making HRIM reusable among many hardware components purposed for robotics. 

\section{Conclusion and future work}
\label{conclusions}

%The landscape of robotics today is dominated by vertical integration where single vendors develop the final product leading to slow developments, expensive products and customer lock-in. Opposite to this, an horizontal integration would result in a rapid development of cost-effective mass-market products with an additional consumer empowerment. The transition of an industry from vertical integration to horizontal integration is typically catalysed by de facto industry standards that enable a simplified and seamless integration of products. However, in robotics there is currently no leading candidate for a global plug-and-play standard.

It is obvious that the robotics community is well aware of the need for standardization, however there is no leading candidate for a global plug-and-play standard. The efforts made so far are country-specific or lack of relevance in simplifying the day to day building robot process. HRIM offers to the robotics community a common interface that facilitates the manufacturing of reusable and interoperable robot hardware modules for the construction of modular robots. An information model for modular robots built upon the principles and abstractions of the popular Robot Operating System (ROS) framework.\\

The contribution of the paper is twofold. First, we document and discuss the current state of the art of models and ontologies in robotics from the perspective of hardware. Second, we present HRIM, an information model for robot hardware, while including examples that explain our design choices.\\

Unlike other models and standardization initiatives, HRIM is focused on hardware. It is being built side by side with manufacturers and experts who actively contribute with their opinions and feedback. Given the continuous technological advances in the robotics industry, HRIM proposes a model that will evolve hand in hand with the available technology. The work presented here is accessible and documented at \url{http://therobotmodel.com}.\\

%Similar to ReApp, HRIM will use a \emph{HRIM components model creation tool} to simplify the implementation. The module manufacturer will choose by an online platform the characteristics and the values that the component has to contain and the package of the specific model will be created automatically.\\

Our team is actively working on HRIM, standardizing a vast variety of components in order to create a solid data base that will increase the choices of interoperable components when designing a robot. Similar to ReApp, we contemplate implementing MDE techniques to make HRIM usable among robotics frameworks, not only with ROS. For this, we should take a step back creating a PIM (Platform Independet Model), making the model framework agnostic. Furthermore, our team would like to explore the possibility of creating an HRIM electronic datasheet, where all the component types could be listed and describe their functions and capabilities electronically. This would simplify the access to the model, spread its information in the community and facilitate the manufacturers' adoption.

We hope for wide acceptance of HRIM, which will lead to better integration of further components and user experience in robotics.\\

\section*{Acknowledgment}
The authors would like to thank Lander Usategui and Asier Bilbao for their contributions supporting the setup and validation of the information model with practical examples on real hardware. The authors would also like to thank Geoffrey Biggs for his feedback and insights on the information model.

%\ifCLASSOPTIONcaptionsoff
%  \newpage
%\fi



%\bibliographystyle{IEEEtran}

\bibliographystyle{IEEEtran}
\bibliography{references}

\end{document}